\documentclass[sigconf]{acmart}
\AtBeginDocument{%
  \providecommand\BibTeX{{%
    \normalfont B\kern-0.5em{\scshape i\kern-0.25em b}\kern-0.8em\TeX}}}

\setcopyright{acmcopyright}
\copyrightyear{2018}
\acmYear{2018}
\acmDOI{10.1145/1122445.1122456}

\acmConference[Woodstock '18]{Woodstock '18: ACM Symposium on Neural
  Gaze Detection}{June 03--05, 2018}{Woodstock, NY}
\acmBooktitle{Woodstock '18: ACM Symposium on Neural Gaze Detection,
  June 03--05, 2018, Woodstock, NY}
\acmPrice{15.00}
\acmISBN{978-1-4503-XXXX-X/18/06}

\usepackage{enumitem} 

\usepackage{amsmath,amsfonts,bm}

\def\eqref#1{equation~\ref{#1}}

\def\1{\bm{1}}

\def\vtheta{{\bm{\theta}}}

\def\vx{{\bm{x}}}
\def\vy{{\bm{y}}}

\DeclareMathAlphabet{\mathsfit}{\encodingdefault}{\sfdefault}{m}{sl}
\SetMathAlphabet{\mathsfit}{bold}{\encodingdefault}{\sfdefault}{bx}{n}

\usepackage{subfigure}
\usepackage{caption}
\usepackage{graphicx}
\usepackage{tabularx}
\usepackage{booktabs}
\usepackage{comment}
\usepackage{wrapfig}
\usepackage{caption}
\usepackage{algorithm}
\usepackage{algorithmicx, algpseudocode}
\usepackage{mathtools}
\usepackage{titlesec}
\usepackage{multirow}
\usepackage{float}
\usepackage{pbox}

\begin{document}

\title{Predictions For Pre-training Language Models}

\author{Tong Guo}
\email{779222056@qq.com}
\affiliation{%
   \country{China}
}

\renewcommand{\shortauthors}{Trovato and Tobin, et al.}

\begin{abstract}
  Language model pre-training has proven to be useful in many language understanding tasks. In this paper, we investigate whether it is still helpful to add the self-training method in the pre-training step and the fine-tuning step. Towards this goal, we propose a learning framework that making best use of the unlabel data on the low-resource and high-resource labeled dataset. In industry NLP applications, we have large amounts of data produced by users or customers. Our learning framework is based on this large amounts of unlabel data. First, We use the model fine-tuned on manually labeled dataset to predict pseudo labels for the user-generated unlabeled data. Then we use the pseudo labels to supervise the task-specific training on the large amounts of user-generated data. We consider this task-specific training step on pseudo labels as a pre-training step for the next fine-tuning step. At last, we fine-tune on the manually labeled dataset upon the pre-trained model. In this work, we first empirically show that our method is able to solidly improve the performance by 3.6\%, when the manually labeled fine-tuning dataset is relatively small. Then we also show that our method still is able to improve the performance further by 0.2\%, when the manually labeled fine-tuning dataset is relatively large enough. We argue that our method make the best use of the unlabel data, which is superior to either pre-training or self-training alone. 
\end{abstract}



\maketitle

\section{Introduction}

\begin{figure*}[t]
	\centering\includegraphics[width=0.8\textwidth,height=0.35\textwidth]{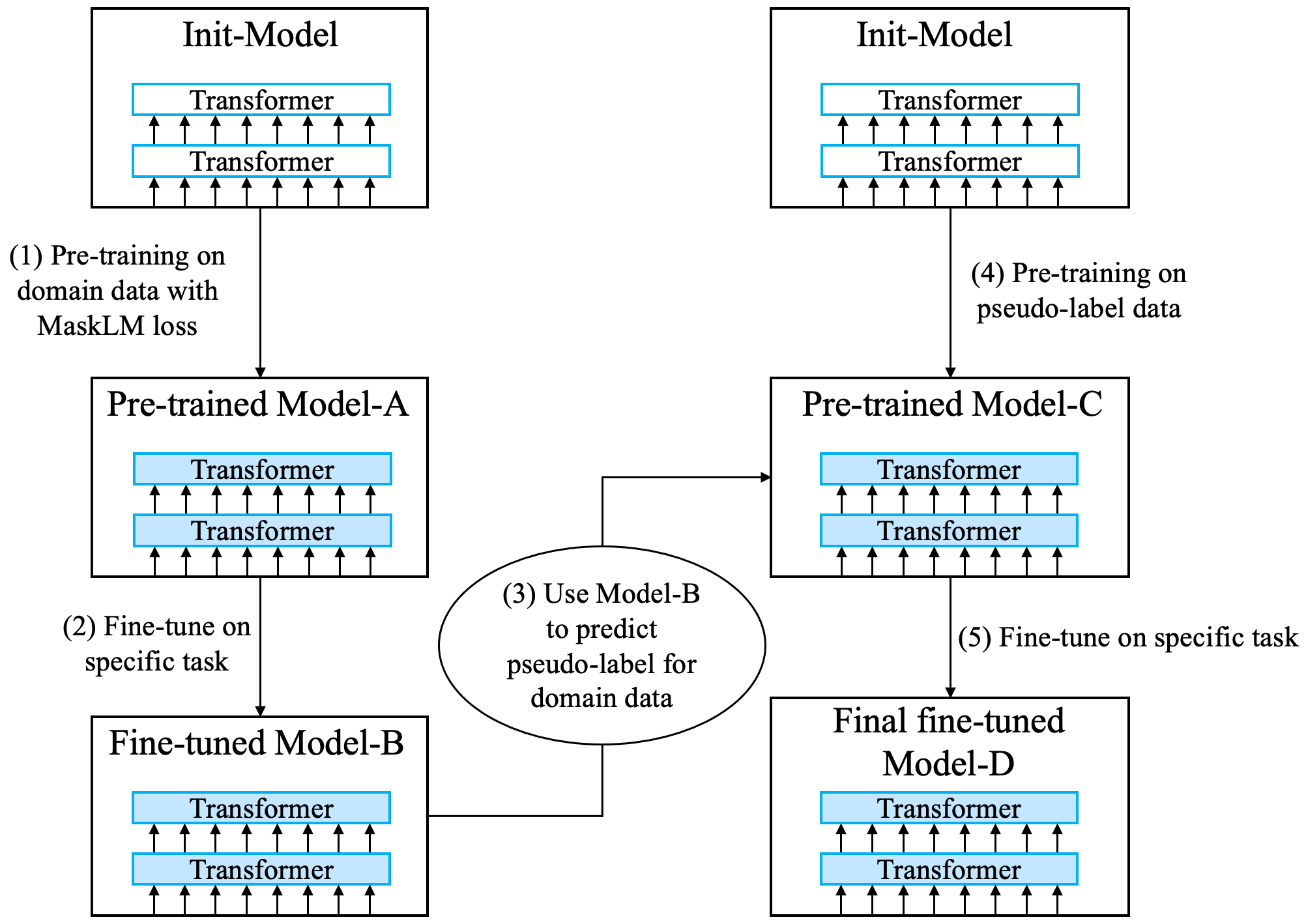}
	\caption{The learning framework of our method: (1) We use the domain-specific data to pre-train on MaskLM task and get the pre-trained Model-A. This step is a classic pre-training step. (2) We use the manually labeled data to fine-tune Model-A to get Model-B. This step is a classic fine-tuning step (3) We use Model-B to predict pseudo labels for all the unlabeled data. (4) We train the model using the pseudo-label and task-specific loss to get Model-C, this step is also considered as a pre-training step for the next step. This step is viewed as the step of self-training as pre-training (5) We fine-tune Model-C to get the final Model-D. The fine-tuning data do not mix the high-score pseudo-label data.}
\end{figure*}

Deep neural networks often require large amounts of labeled data to achieve good performance. However, acquiring labels is a costly process, which motivates research on methods that can effectively utilize unlabeled data to improve performance. Towards this goal, semi-supervised learning \cite{chapelle2009semi} and pre-training are proposed to take advantage of both labeled and unlabeled data. Self-training \cite{scudder1965probability,yarowsky1995unsupervised} is a semi-supervised method which uses a teacher model, trained using labeled data, to create pseudo labels for unlabeled data. Then, a student model is trained with this new training set to yield the final model. Meanwhile, language model pre-training has been shown to be effective for improving many natural language processing (NLP) tasks \cite{devlin2019bert,brown2020language,peters2018deep,radford2019language,radford2018improving}. Previous studies \cite{gururangan2020don,lee2020biobert,beltagy2019scibert} also have shown the benefit of continued pre-training on domain-specific unlabeled data. Then our direct motivation is to answer the question: Do pre-training combined with self-training further improves the fine-tuning performance? Are they complementary? In this paper, we try to answer this question by using the pseudo-label data for pre-training language model.

We are one of the largest Chinese dining and delivery platform. The report shows that the total number of online food orders reaches millions in a day. In our industry NLP applications, we gather large amounts of unlabeled data produced by users or customers. So we have large amounts of data in the food domain, which allows us to use the large amounts of unlabel data. We follow the self-training idea that we can use the fine-tuned model to predict a pseudo-label. We explore the question whether the data with pseudo-label bring further improvement for BERT. We design a learning framework to explore the effect of self-training for pre-training. The whole learning framework is shown in Figure 1. 

In this paper, we evaluate the performance of our learning framework on our item (a specific food) type text classification task and item property named entity recognition (NER) task. For the item type text classification task, our goal is to predict the 1 food type in 311 classes/types for all the 500 million items, given the item name, item short tag written by POI (Point of Interest, a specific store or restaurant) and item's POI name. For the item property NER task, our goal is to extract item property (such as benefit for stomach, sweet and chewy, cool and refreshing) from item description. The task detail is described in section 3.

As it is shown in Figure 1, we have several terminologies to introduce:

\textbf{In-domain data / domain-specific data}: The large amounts of data in our database. Our manually labeled dataset is sampled from this in-domain data. This in-domain data is produced by our user or customer.

\textbf{Self-training}: We use a simple self-training method inspired by Algorithm 1. First, a teacher model is trained on the manually labeled data. Then the teacher model generates pseudo labels on unlabeled data (e.i., all the data in our database). Finally, a student is trained to optimize the loss on human labels and pseudo labels jointly.

\textbf{Pre-training}: This term means the training step that use hundred millions data. The step 1 of Figure 1 is training on the unlabel data. The step 1 of Figure 1 is training on the pseudo-label data. The next step of this training step is the fine-tuning step using the manually labeled data.

\textbf{Self-training as pre-training / task-specific training}: This term refers to the step 4 of Figure 1, which means the training step that use pseudo-label. We consider this training step as a pre-training step for the next fine-tuning step on manually labeled data. 

\textbf{Self-training for fine-tuning / task-specific fine-tuning}: This term refers to the step 5 of Figure 1, which means the fine-tuning step that use both the pseudo-label data and the manually labeled data. 

In detail, we aim to answer the questions: How much does self-training as pre-training perform in low-resource NLP task and high-resource NLP task? To the best of our knowledge, this is the first study exploring the improvement of self-training as pre-training (i.e., using the pseudo-label data for pre-training language models) for natural language understanding. And this is the first study exploring the self-training improvement based on the step of self-training as pre-training (i.e., adding the high-confidence-score pseudo-label data for fine-tuning based on the pre-trained model). This is also the first work exploring the combination of self-training and pre-training when the manually labeled fine-tuning dataset is relatively large (2000K).

In summary, our contributions include:

• We explore the improvement of pre-training combined with self-training. We reveal that pre-training combined with self-training improves the performance stably, when the fine-tuning dataset is relatively small (100K). We find that using the pseudo-label data for fine-tuning do not improve the performance further, when the fine-tuning dataset is relatively large enough. But using the manually labeled data in the fine-tuning step without pseudo-label data improve the performance, when the fine-tuning dataset is relatively large enough (2000K).

• We argue that our learning framework is the best combination of self-training and pre-training method to make use of a large amounts of unlabeled data. Even when the fine-tuning dataset is relatively large enough (2000K), Our method, which is corresponding to Figure 1 and Figure 4-5, still is able to improve the performance. In our experiments based on our dataset, the previous methods (i.e., the classic self-training) is not able to improve the performance when the fine-tuning dataset is relatively large enough (2000K).

• We explore different experimental factors on our text classification dataset and NER dataset. The experiment results prove that the pre-training with task-specific loss and no-MaskLM (masked language model) loss is the best way to make use of the unlabel data. We also find that pre-training using KL-divergence loss with pre-softmax logits is better than cross-entropy loss with one-hot pseudo-label, which is corresponding to the step 4 of Figure 1.

\section{Related Work}

\begin{algorithm}[!t]
    \centering
    \caption{Classic Self-training}
    \label{alg:st}
    \begin{algorithmic}[1]
    \State Train a base model $f_{\vtheta}$ on $L=\{\vx_i, \vy_i\}_{i=1}^l$ 
    \Repeat
        \State Apply $f_{\vtheta}$ to the unlabeled instances $U$
        \State Select a subset $S \subset \{(\vx, f_{\vtheta}(\vx)) | \vx \in U\}$
        \State Train a new model $f_{\vtheta}$ on $S \cup L$
    \Until{convergence or maximum iterations are reached}
    \end{algorithmic}
    \end{algorithm}

There is a long history of pre-training language representations\cite{brown1992class,ando2005framework,blitzer2006domain,pennington2014glove,mikolov2013distributed,turian2010word,mnih2009scalable,kiros2015skip,logeswaran2018efficient,jernite2017discourse,hill2016learning}, and we briefly review the most widely-used approaches in this section.

BERT \cite{devlin2019bert} is based on the multi-layer transformer encoder \cite{vaswani2017attention}, which produces contextual token representations that have been pre-trained from unlabeled text and fine-tuned for a supervised downstream task. BERT achieved previously state-of-the-art results on many sentence-level tasks from the GLUE benchmark \cite{wang2018glue}. There are two steps in BERT's framework: pre-training and fine-tuning. During pre-training, the model is trained on unlabeled data by using masked language model task and next sentence prediction task. Apart from output layers, the same architectures are used in both pre-training and fine-tuning. The same pre-trained model parameters are used to initialize models for different down-stream tasks.

Semi-supervised learning \cite{zhu2009introduction,zhu2005semi,chapelle2009semi} solve the problem that making best use of a large amounts of unlabeled data. These works include UDA \cite{xie2020unsupervised}, Mixmatch \cite{berthelot2019mixmatch}, Fixmatch \cite{sohn2020fixmatch}, Remixmatch \cite{berthelot2019remixmatch}. These works design the unsupervised loss to add to the supervised loss. These works prove that domain-specific unlabel data is able to improve the performance, especially in low-resource manually labeled dataset.

Self-training \cite{blum1998combining,zhou2004democratic,zhou2005tri} is one of the earliest and simplest semi-supervised methods. As shown in Algorithm 1, Self-training first uses labeled data to train a good teacher model, then use the teacher model to label unlabeled data and finally use the labeled data and unlabeled data to jointly train a student model. Some early work have successfully applied self-training to word sense disambiguation\cite{yarowsky1995unsupervised} and parsing\cite{huang2009self,reichart2007self,mcclosky2006effective}. In recent years, these works include self-training for natural language processing\cite{du2020self,he2019revisiting}, self-training for computer vision \cite{xie2020self,zoph2020rethinking}, self-training for speech recognition \cite{kahn2020self} and back-translation for machine translation \cite{bojar2011improving,sennrich2015improving,edunov2018understanding}.  \cite{zoph2020rethinking} reveals some generality and flexibility of self-training combined with pre-training in computer vision. \cite{du2020self} study the self-training improvement on the fine-tuning step, based on data augmentation. But \cite{du2020self} restrict to the size of the specific in-domain data and manually labeled data.


\section{Our Dataset}

In this section, we describe our datasets. The item examples are shown in Table 1. The task examples are shown in Table 2. The data size information is shown in Table 3.

\begin{table*}[t]
	\centering
	\def\arraystretch{2}
	\caption{Some examples of the food/items. }
	
	\begin{tabular}{cccc}
		 \toprule
         Item Name & Item Short Tag  & Item POI Name  & Item Description \\\hline
         
		Yuxiang shredded pork & Hot dishes  & The Sichuan restaurant &  \pbox{6cm}{It contains salty, sweet, sour, hot, and fresh tastes,  making the food more delicious.}\\
	    Kung Pao chicken & Hot dishes  & The Sichuan restaurant &  \pbox{6cm}{It is a spicy, stir-fried Chinese dish made with cubes of chicken, peanuts and vegetables.}\\
		Congee with Pork and Century Eggs & Congee series  & The congee restaurant &  \pbox{6cm}{It is benefit for stomach.}\\
		Knife-sliced noodle & Noodle series & The noodle restaurant &  \pbox{6cm}{The noodles going directly into the boiling water. This makes the noodles very fresh}\\
        
		\bottomrule
	\end{tabular}
	
\end{table*}

\begin{table*}[t]
	\centering
	\caption{Some examples of the text classification task and the NER task, corresponding to Table 1.}
	
	\begin{tabular}{ccc}
		 \toprule
         Item Name & Item Classification Food Type  & Item NER results  \\\hline
         
		Yuxiang shredded pork & stir-fried food  &  fresh tastes \\
	    Kung Pao chicken & stir-fried food  & (no result) \\
		Congee with Pork and Century Eggs & meat congee  & benefit for stomach \\
		Knife-sliced noodle &  noodle with soup & very fresh \\
        
		\bottomrule
	\end{tabular}
	
\end{table*}

\begin{table*}[t]
	\centering
	
	\caption{The data amount and average token number for each experiment step. }
	
	\begin{tabular}{c|cc|cc}
		 \toprule
		&\multicolumn{2}{c|}{Text Classification Task}&\multicolumn{2}{c}{NER Task}\\
        Experiment Step &   Data Size & Average Length  & Data Size & Average Length \\\hline
		Domain-Specific Pre-training & 500 million & 28  & 200 million & 50 \\
        Task-Specific Pre-training & 500 million & 28  & 200 million & 50\\
		Fine-Tune Training & 2,000,000& 28 & 47,500 & 50\\
		Fine-Tune Test & 40,000 & 28 & 2,500 & 50\\
		\bottomrule
	\end{tabular}
	
\end{table*}

\subsection{Chinese Text Classification Dataset}

This task is to predict the item type or category, given the item name, item short tag, item POI name. We define 311 classes/types for all the items. The model inputs are item name, item short tag given by POI and the POI name of item. We manually labeled 2,040,000 data. The total item number is 500 million.

\subsection{Chinese NER Dataset}

This task is to extract all the item properties from item description. The item description is a short paragraph written by users. We manually labeled 50,000 data. The total item description number is 200 million. There are 500 million items in total, in which 200 million items have the their descriptions.

\section{Our Method}

In this section, we describe our method in detail. As the Figure 1 shows, our framework includes 5 steps, so we separate this section into 5 subsections: In subsection 4-1, we describe the domain-specific pre-training with unlabeled data. In subsection 4-2, we describe the task-specific fine-tuning with the manually labeled data. In subsection 4-3, we describe the inference step by the fine-tuned model of last step.  In subsection 4-4, we describe the task-specific pre-training with the pseudo-label predicted by the fine-tuned model. In this paper, the step that training the model with the task-specific loss on pseudo-label data is also considered as a pre-training step for the next step. In subsection 4-5, we describe the task-specific fine-tuning which is almost same to subsection 4-2 to get the final model.

\subsection{Domain-Specific Pre-training}
This is the first step of our method. This step is almost the same to the origin BERT's \cite{devlin2019bert} pre-training step except the data preparation. We use all the in-domain data in our database for pre-training.  

For our Chinese text classification task, the model's 3 inputs are item name, item short tag and item's POI name. We follow the origin BERT \cite{devlin2019bert} setting and use the character-level word masking. In detail we concat the 3 inputs string and mask 15\% characters. The max sequence length is 64. We follow the RoBERTa \cite{liu2019roberta} setting and remove the next-sentence-predict loss. For efficiency reason, we use 3-layer-BERT in our experiment because we need to inference more than hundred millions data in our application. We extract 3 layers from the origin official pre-trained BERT \cite{devlin2019bert} as the initialized parameter. The total pre-training data number is 500 million. 

For our Chinese NER task, the model's input is the item description short paragraph. We follow the RoBERTa setting and use the character-level word masking in the short paragraph. The max sequence length is 128 and we remove the next-sentence-predict loss. For efficiency reason, we use 3-layer-BERT in our experiment. We extract 3 layers from the origin official pre-trained BERT as the initialized parameter. The total pre-training data number is 200 million.

\begin{figure}[t]
	\centering\includegraphics[width=0.35\textwidth,height=0.24\textwidth]{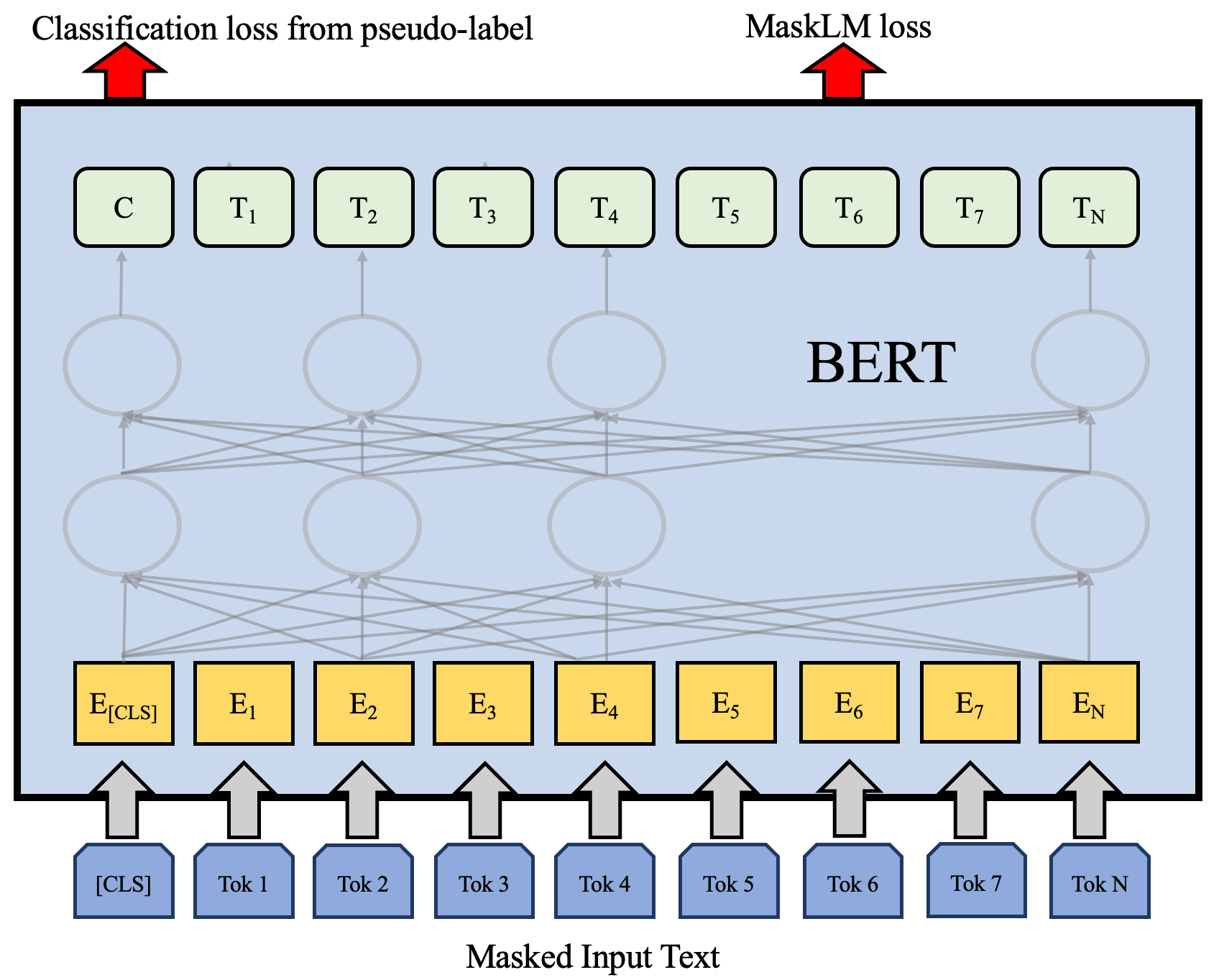}
	\caption{The task-specific pre-training for item type text classification task (i.e., the ablation experiment for step 4 of Figure 1). The main difference to the origin BERT pre-training is we replace the next-sentence-predict loss with the classification loss from the pseudo-label. In this picture, we mask 15\% characters of the text as the model input.}
\end{figure}

\begin{figure}[t]
	\centering\includegraphics[width=0.35\textwidth,height=0.24\textwidth]{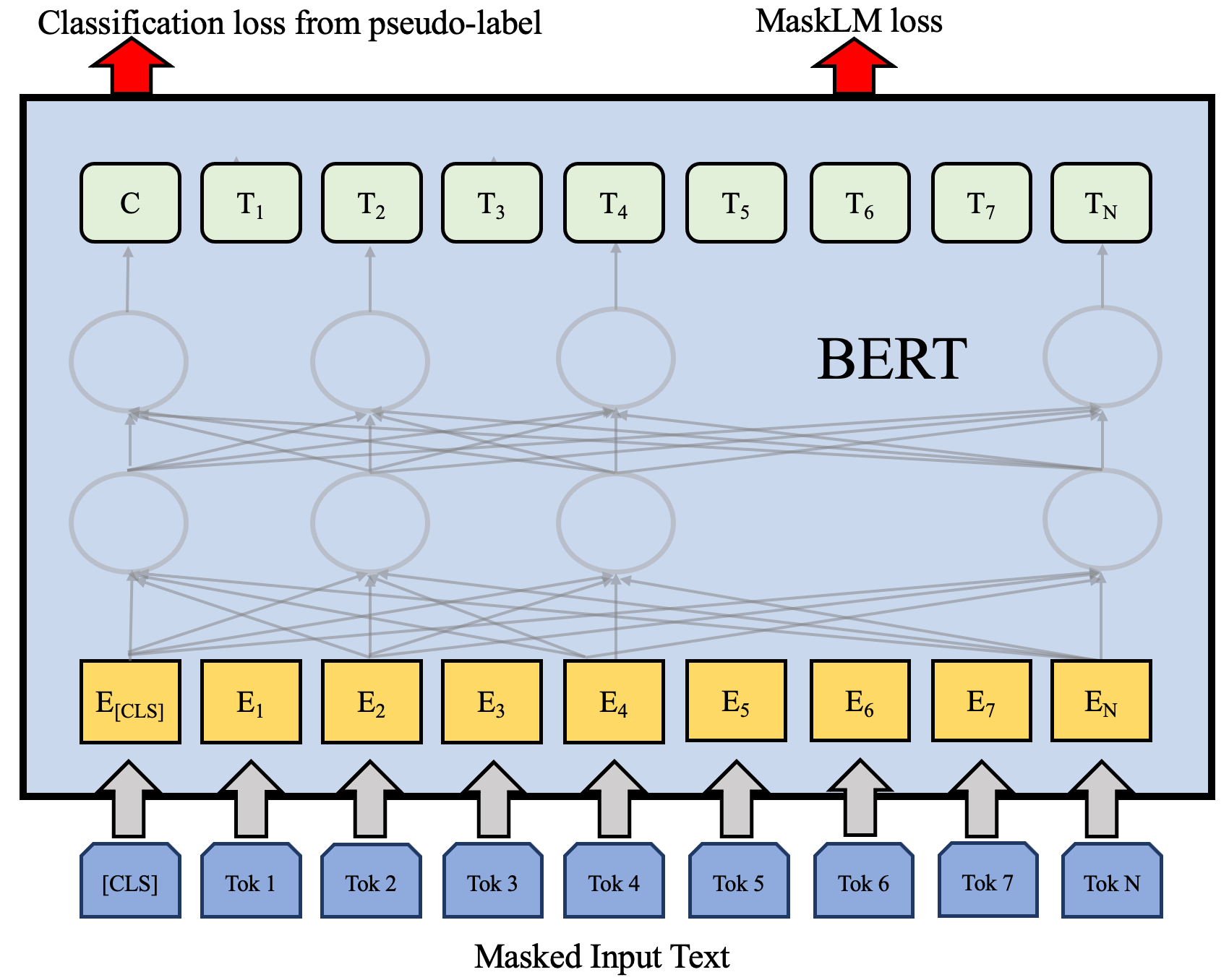}
	\caption{The task-specific pre-training for item property NER task (i.e., the ablation experiment for step 4 of Figure 1). The main difference to the origin BERT pre-training is we add the NER loss from the pseudo-label. In this picture, we mask 15\% characters of the text as the model input.}
\end{figure}

\begin{figure}[t]
	\centering\includegraphics[width=0.35\textwidth,height=0.24\textwidth]{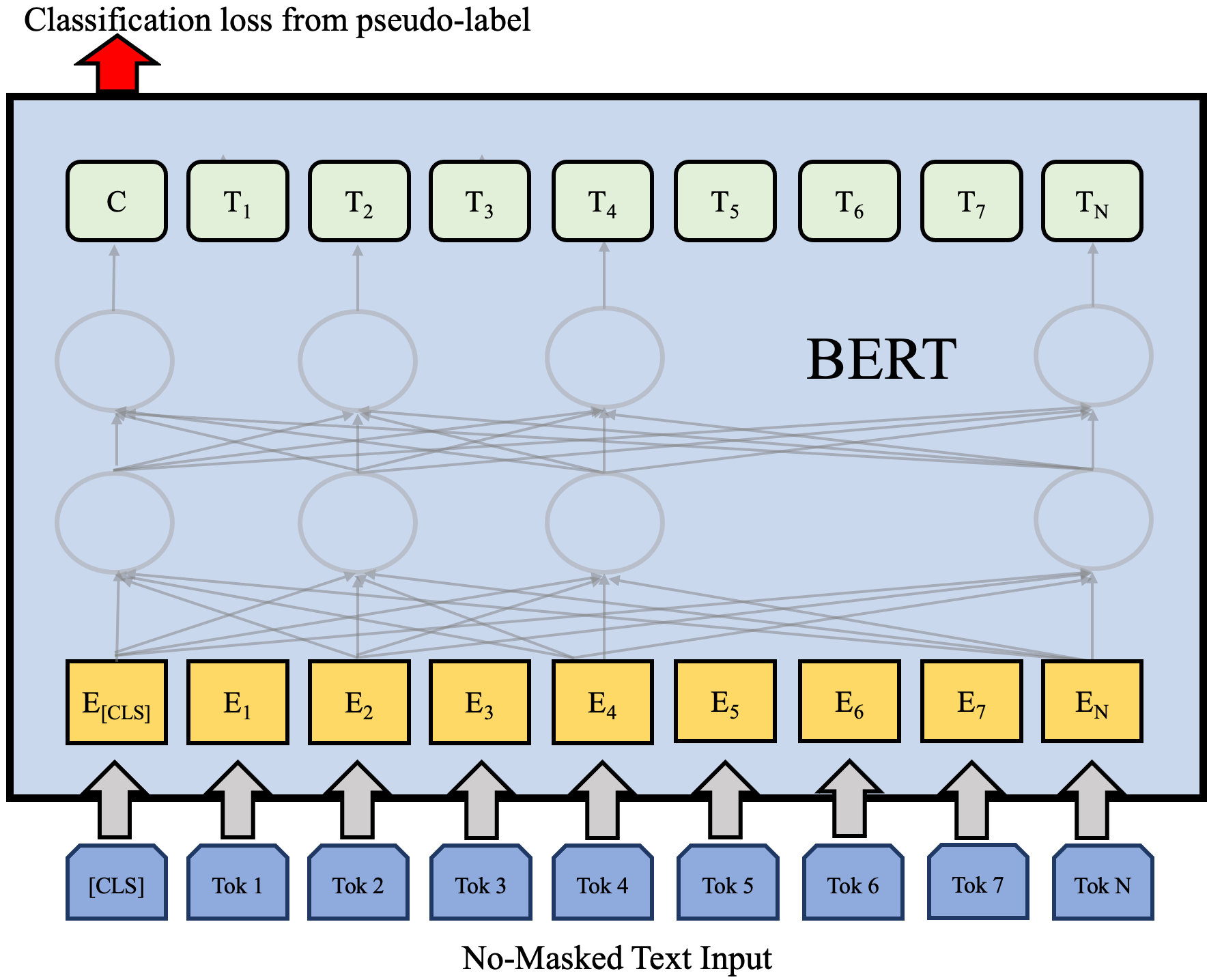}
	\caption{The task-specific pre-training for item type text classification task (i.e., step 4 of Figure 1). The main difference to the origin BERT pre-training is we replace the next-sentence-predict loss with the classification loss from the pseudo-label. In this picture, we do not mask 15\% characters of the text as the model input.}
\end{figure}

\begin{figure}[t]
	\centering\includegraphics[width=0.35\textwidth,height=0.24\textwidth]{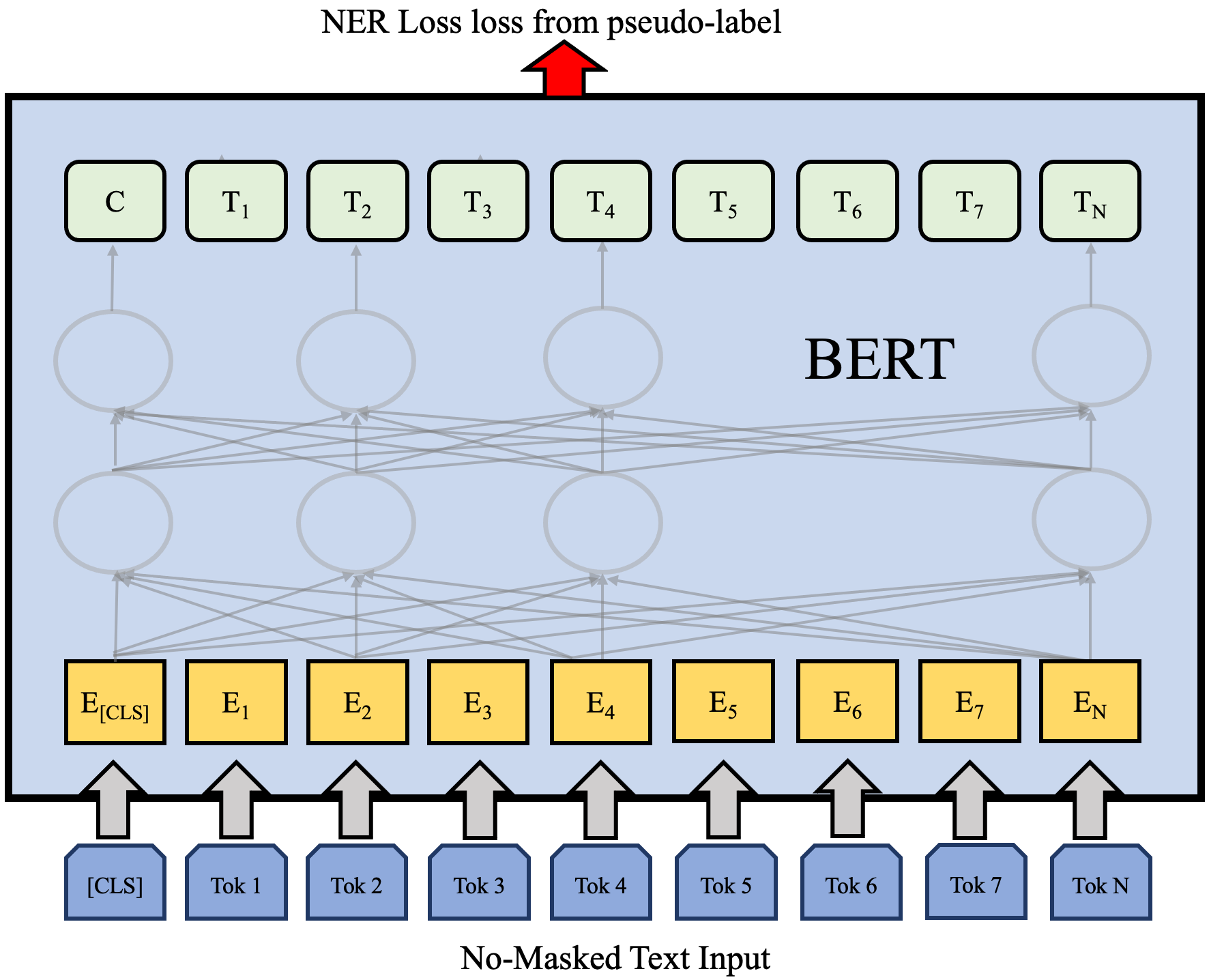}
	\caption{The task-specific pre-training for item property NER task (i.e., step 4 of Figure 1). The main difference to the origin BERT pre-training is we replace the MaskLM loss with the NER loss from the pseudo-label. In this picture, we do not mask 15\% characters of the text as the model input.}
\end{figure}

\subsection{Task-Specific Fine-Tuning}

This is the second step of our method. We use the model parameter of the last step to initialize this step's model. 

For our Chinese text classification task, we follow the origin BERT \cite{devlin2019bert} setting and use the cross-entropy loss. The total fine-tuning data number is our 2,040,000 manually labeled data. We select 40,000 as test dataset. Then we sampled 100K, 400K and 2000K data from the rest 2000K data separately.

For our Chinese NER task, we use the CRF(Conditional Random Field) loss. The total fine-tuning data number is our 50,000 manually labeled data. We split all the data to 19:1 as training dataset and test dataset.

\subsection{Inference Step}

This is the third step of our method. This step is to use the fine-tuned model of last subsection to predict a pseudo-label for all the in-domain data.

For our Chinese text classification task, we use the fine-tuned model of the last subsection to predict the classification pseudo-label for all the 500 million unlabeled data. Then we get the 500 million classification data with pseudo-label. The pseudo-label could be the one-hot label or the pre-softmax logits.

For our Chinese NER task, we use the fine-tuned model of the last subsection to predict the sequence tagging pseudo-label for all the 200 million unlabeled data. Then we get the 200 million NER data with pseudo-label. The pseudo-label could be the one-hot label sequence or the pre-softmax logits.

\subsection{Task-specific Pre-training}

This is the fourth step of our method. We use the pseudo-label of last step as the pseudo ground truth label for training. We consider this training step as the pre-training step for the next step.

For our Chinese text classification task, the model input is the masked text, which is the concat of item name, item short tag and item's POI name. There are four kinds of experiment: 1) The sum of the classification cross-entropy loss and the MaskLM loss. 2) The cross-entropy loss only. 3) The sum of the classification KL-divergence loss on the pre-softmax logits and the MaskLM loss. 4) The KL-divergence loss on the pre-softmax logits only. In detail, we use the [CLS] token's output of BERT for computing the classification cross-entropy loss on one-hot pseudo-label or the KL-divergence loss on the pre-softmax logits. The detail is shown in Figure 2 and Figure 4.

For our Chinese NER task, the model input is masked text, which is the item description short paragraph. There are four kinds of experiment: 1) The sum of the CRF loss and the MaskLM loss. 2) The CRF loss only. 3) The sum of the KL-divergence loss on the pre-softmax logits and the MaskLM loss. 4) The KL-divergence loss on the pre-softmax logits only. In detail, we use all the tokens' output sequence of BERT for computing the CRF loss and MaskLM loss. The detail is shown in Figure 3 and Figure 5.

\subsection{Final Task-specific Fine-tuning}

This is the final step of our method. We find that fine-tuning on the manually labeled dataset without pseudo-label high-score data is better. For the classic self-training comparison experiment, we randomly sampled the addition data from each class of the pseudo-label high-score data. We use the pre-trained model of subsection 4-4 for fine-tuning.

\begin{table*}[t]
	\centering
	 
	\caption{Experiment results for the NER task and the text classification task separated to fine-tuning data with different size. The BERT-Base baselines are initialized by the origin BERT's \cite{devlin2019bert}. Classic self-training follow the Algorithm 1 and mix the pseudo-label data and manually labeled data. Our method is pre-training using KL-divergence loss on the pre-softmax logits without the MaskLM loss and fine-tuning on dataset that do not mix the high-score pseudo-label data, which is corresponding to Figure 1. Accuracy scores are reported for the text classification task. F1 scores are reported for NER task. }

	\begin{tabular}{c|c|c|c|c}
	\toprule
		&\multicolumn{3}{c|}{Text Classification Test Accuracy}&\multicolumn{1}{c}{NER}\\
		Model &  100K data & 400K data & 2000K data  &  Test F1 \\\hline
		BERT-Base-3layer baseline & 87.0 & 89.2 & 90.8   & 88.6 \\
		BERT-Base-12layer baseline & 87.5 & 89.4 & 91.0   & 88.7 \\
		Classic Self-training upon BERT-Base-3layer & 88.0 & 89.8 & 91.0   & 89.0 \\
		Model-B & 88.2 & 90.1 & 91.7  & 88.8 \\
		Classic Self-training upon Model-A & 89.3 & 90.4 & 91.6   & 89.2 \\
		Classic Self-training upon Model-C & 90.2 & 90.9 & 91.7 &  89.4 \\
		Our Method (Model-D) & \textbf{90.6} & \textbf{91.1} &  \textbf{91.9} &  \textbf{89.6} \\
	\bottomrule
	\end{tabular}
	
\end{table*}

\section{Experiments}
In this section we describe our experimental setup and experimental results.

\subsection{Experimental Setup}

In this section we describe the parameters in our experiments. In all the experiments except the baselines, we use 3-layer-BERT with 768 hidden size and 12 self-attention heads. (Total Parameters = 40M). For efficiency reason, we use only 3 layers of BERT, because it is an industry application and we only have limited computation resource to inference hundred millions data. And 3-layer-BERT is 200\%-300\% faster than 12-layer-BERT in inference time. The max sequence length for text classification task is 64. The max sequence length for the NER task is 128. The total data size are presented in Table 3. For the text classification task, we setup 3 experimental groups with different fine-tuning data size separately. The 3 different fine-tuning dataset are sampled from the total 2,000,000 data and run the learning framework independently. 

\subsubsection{Text Classification Domain-Specific Pre-training}

We use a batch size of 64 * 4-GPU and pre-train for 3,000,000 steps. We use Adam \cite{kingma2014adam} with learning rate of 5e-5, beta1 = 0.9, beta2 = 0.999, L2 weight decay of 0.01, learning rate warmup over the first 10,000 steps, and linear decay of the learning rate. We use a dropout \cite{srivastava2014dropout} probability of 0.1 on all layers. The 3-layer-BERT is initialized by the origin BERT's \cite{devlin2019bert} 3 layers. 

\subsubsection{NER Domain-Specific Pre-training}
The NER domain-specific pre-training is almost the same to the text classification domain-specific pre-training. The pre-training data is item description sentence for NER and the pre-training data is concatenation of multi words (item name, item short tag, item poi name) for text classification.

\subsubsection{Text Classification Fine-Tuning}
We use a batch size of 64 * 1-GPU and fine-tune for 7 epochs. We use Adam with learning rate of 1e-5. The dropout probability is 0.1 on all layers. The fine-tuning data with different size in Table 2 is sampled from all the 2,000,000 data and the 40,000 test data is fixed.

\subsubsection{NER Fine-Tuning}
We use a batch size of 64 * 1-GPU and fine-tune for 3 epochs. We use Adam with learning rate of 1e-5. The dropout probability is 0.1 on all layers.

\subsubsection{Text Classification Task-Specific Pre-training}
We use a batch size of 64 * 4-GPU and pre-train for 3,000,000 steps. We use Adam with learning rate of 5e-5, beta1 = 0.9, beta2 = 0.999, L2 weight decay of 0.01, learning rate warmup over the first 10,000 steps, and linear decay of the learning rate. We use a dropout probability of 0.1 on all layers. The 3-layer-BERT is initialized by the origin BERT's \cite{devlin2019bert} 3 layers. For fair comparison, we do not initialize this step of pre-training by the result model of domain-specific pre-training step.

\subsubsection{NER Task-Specific Pre-training}
We use a batch size of 64 * 4-GPU and pre-train for 3,000,000 steps. We use Adam with learning rate of 5e-5, beta1 = 0.9, beta2 = 0.999, L2 weight decay of 0.01, learning rate warmup over the first 10,000 steps, and linear decay of the learning rate. We use a dropout probability of 0.1 on all layers. The 3-layer-BERT is initialized by the origin BERT's \cite{devlin2019bert} 3 layers. For fair comparison, we do not initialize this step of pre-training by the result model of domain-specific pre-training step.

\begin{table*}[t]
	\centering
	 
	\caption{Ablation over the pre-training step on in-domain data without task-specific loss, corresponding to the step 1 of Figure 1, which shows the improvement of in-domain pre-training on MaskLM task. Model-B is corresponding to Figure 1.}

	\begin{tabular}{c|c|c|c|c}
	\toprule
		&\multicolumn{3}{c|}{Text Classification Test Accuracy}&\multicolumn{1}{c}{NER}\\
		Model &  100K data & 400K data & 2000K data  &  Test F1 \\\hline
		BERT-Base-3layer baseline & 87.0 & 89.2 & 90.8   & 88.6 \\
		BERT-Base-12layer baseline & 87.5 & 89.4 & 91.0   & 88.7 \\
		Model-B & \textbf{88.2} & \textbf{90.1} & \textbf{91.7}  & \textbf{88.8} \\
	\bottomrule
	\end{tabular}
	
\end{table*}

\begin{table*}[t]
	\centering
	 
	\caption{Ablation over the pre-training step on in-domain unlabel data and pseudo-label data, which shows the improvement of pre-training on pseudo-label data. Model-B and Model-D are corresponding to Figure 1. Model-B is upon the in-domain MaskLM-based pre-training on unlabel data. Model-D is upon the task-specific pre-training on pseudo-label data.}

	\begin{tabular}{c|c|c|c|c}
	\toprule
		&\multicolumn{3}{c|}{Text Classification Test Accuracy}&\multicolumn{1}{c}{NER}\\
		Model &  100K data & 400K data & 2000K data  &  Test F1 \\\hline
		Model-B & 88.2 & 90.1 & 91.7  & 88.8 \\
		Model-D & \textbf{90.6} & \textbf{91.1} &  \textbf{91.9} &   \textbf{89.6} \\
	\bottomrule
	\end{tabular}
	
\end{table*}

\begin{table*}[t]
	\centering
	 
	\caption{Ablation over the task-specific pre-training step between logits-based loss and one-hot label-based loss, corresponding to the step 4 of Figure 1. Model-D is corresponding to Figure 1. (label) means we use the one-hot pseudo-label for the supervised pre-training. (logits) means we use the pre-softmax logits for the supervised pre-training.}

	\begin{tabular}{c|c|c|c|c}
	\toprule
		&\multicolumn{3}{c|}{Text Classification Test Accuracy}&\multicolumn{1}{c}{NER}\\
		Model &  100K data & 400K data & 2,000K data  &  Test F1 \\\hline
		Model-D (label) & 90.4 & 91.0 & 91.7 &  89.3 \\
		Model-D (logits) & \textbf{90.6} & \textbf{91.1} &  \textbf{91.9} &  \textbf{89.6} \\
	\bottomrule
	\end{tabular}
	
\end{table*}

\begin{table*}[t]
	\centering
	 
	\caption{Ablation over the task-specific pre-training step between masked noisy input text and and the origin input text, corresponding to the step 4 of Figure 1. Model-D is corresponding to Figure 1. (masked) means we masked 15\% tokens of the input text. (nomask) means we input the origin text.}

	\begin{tabular}{c|c|c|c|c}
	\toprule
		&\multicolumn{3}{c|}{Text Classification Test Accuracy}&\multicolumn{1}{c}{NER}\\
		Model &  100K data & 400K data & 2000K data  &  Test F1 \\\hline
		Model-D (masked / MaskLM loss + Task-specific loss) & 90.2 & 90.9 &  91.6 &  89.3 \\
		Model-D (nomask / Task-specific loss) & \textbf{90.6} & \textbf{91.1} &  \textbf{91.9} &   \textbf{89.6} \\
	\bottomrule
	\end{tabular}
	
\end{table*}

\begin{table*}[t]
	\centering
	 
	\caption{The comparison of absolute improvement upon baseline (the first row of table), which proves the conclusion again:  Continued pre-training on in-domain unlabeled data benefits the performance.}

	\begin{tabular}{c|c|c|c|c}
	\toprule
		&\multicolumn{3}{c|}{Text Classification Test Accuracy}&\multicolumn{1}{c}{NER}\\
		Factor &  100K data & 400K data & 2000K data  &  Test F1 \\\hline
		BERT-Base-3layer baseline & - & - &  - &  - \\
	    BERT-Base-12layer & +0.5\% & +0.2\% & +0.2\%  & +0.1\%  \\
	    + self-training & +1.0\% & +0.6\% & +0.2\%  & +0.4\%  \\
	    In-domain pre-training with MaskLM task & +1.2\% & +0.9\% & +0.9\%  & +0.2\%  \\
	\bottomrule
	\end{tabular}
	
\end{table*}

\begin{table*}[t]
	\centering
	 
	\caption{The comparison of absolute improvement upon baseline (the first row of table), which reveals the conclusion: Self-training as pre-training get bigger gain than in-domain pre-training.}

	\begin{tabular}{c|c|c|c|c}
	\toprule
		&\multicolumn{3}{c|}{Text Classification Test Accuracy}&\multicolumn{1}{c}{NER}\\
		Factor &  100K data & 400K data & 2000K data  &  Test F1 \\\hline
	    In-domain pre-training with MaskLM task  & - & - &  - &  - \\
	    + self-training & +0.5\% & +0.3\% & -0.1\%  & +0.4\%  \\
	    pre-training with task-specific task and MaskLM task & +1.0\% & +0.8\% & -0.1\%  & +0.5\%  \\
	    pre-training with task-specific task & +1.4\% & +1.1\% & +0.2\%  & +0.8\%  \\
	\bottomrule
	\end{tabular}
	
\end{table*}

\begin{table*}[t]
	\centering
	 
	\caption{The comparison of absolute improvement upon baseline (the first row of table), which reveals the conclusion: For high-resource manually labeled dataset, pre-training with task-specific loss without masking the input text is the only way to further improve the performance.}

	\begin{tabular}{c|c|c|c|c}
	\toprule
		&\multicolumn{3}{c|}{Text Classification Test Accuracy}&\multicolumn{1}{c}{NER}\\
		Factor &  100K data & 400K data & 2000K data  &  Test F1 \\\hline
	    pre-training with task-specific task  & - & - &  - &  - \\
	    + self-training & +0.4\% & +0.2\% & -0.2\%  & +0.2\%  \\
	\bottomrule
	\end{tabular}
	
\end{table*}

\subsection{Baseline Setup}
In this section we describe the baselines in Table 2.
\subsubsection{BERT-Base-3layer baseline} 
 For the BERT-Base-3layer baseline, we extract 3 layers from the origin 12-layer BERT-Base \cite{devlin2019bert} and fine-tune on the manually labeled dataset. 
 \subsubsection{BERT-Base-12layer baseline}
 For the BERT-Base-12layer baseline, we use the origin 12-layer BERT-Base \cite{devlin2019bert} and fine-tune on the manually labeled dataset. 
 \subsubsection{Classic Self-training baseline}
 For the Classic Self-training baseline, we use a simple self-training method inspired by Algorithm 1. First, a teacher model is trained on the manually labeled data. Then the teacher model generates pseudo labels on unlabeled data (e.i., all the data in our database). Finally, a student is trained to optimize the loss on human labels and pseudo labels jointly. The pseudo labels data with the high confidence score are averagely sampled from each class of all the data for fine-tuning upon the origin BERT-Base-12layer, Model-A and Model-C. 

\subsection{Experimental Results}
In this section, we present experiment results on the text classification task and the NER task. The detail results are presented in Table 4-11. The text classification accuracy means the exact match of predicted class and the ground truth. There are 311 classes in the task. The NER F1 is same to the definition of CoNLL \cite{sang2003introduction}.  The BERT-Base baseline's pre-trained model is from the origin official Github repository. The BERT-Base-3layer is extracted from the origin official BERT-Base-12layer.

\subsection{Analysis}

In this section, we analysis our experiment results on Table 4.  we focus on the effect of two factors: the size of the fine-tuning dataset, the data content of the pre-training dataset.

\textbf{Fine-tuning data size.} In low-resource dataset, the whole gain of our framework is 3.6\% (from 87.0\% to 90.6\%) for the text classification task. In high-resource dataset, the whole gain of our framework is 1.1\% (from 90.8\% to 91.9\%) for the text classification task. We also observe that pre-training with pseudo-label data (i.e., step 4 of Figure 1) is able to improve the performance in high-resource dataset while fine-tuning with manually labeled data and pseudo-label data is not able to improve the performance in high-resource dataset. In the NER task, the stable improvement of in-domain pre-training (i.e., step 1 of Figure 1) is only 0.2\% (from 88.6\% to 88.8\%), compared to the whole 1.0\% (from 88.6\% to 89.6\%) improvement of the learning framework in the NER task. While the improvement of in-domain pre-training for the text classification task is 0.9\% (from 90.8\% to 91.9\%). And the whole improvement is 1.1\% (from 90.8\% to 91.9\%) for the text classification task.

\textbf{In-domain data and out-domain data.} The domain-specific pre-training (i.e., the step 1 of Figure 1) gain of the text classification task is more than the gain of the NER task. The gain of in-domain pre-training (i.e., the step 1 of Figure 1) for text classification task is 0.7\% (from 87.5\% to 88.2\%). The gain of in-domain pre-training (i.e., the step 1 of Figure 1) for the NER task is 0.1\% (from 88.7\% to 88.8\%). The reason is that the input of the text classification task is the concatenation of multi words (item name, item short tag, item poi name), which is not consistent to the origin BERT's \cite{devlin2019bert} pre-training. The origin BERT's pre-training uses the Chinese Wikipedia natural language sentences. And the NER task's in-domain pre-training uses natural language sentences in our database, which is not that much different to the data format of origin BERT.

\subsection{Ablation Studies}
In this section, we perform ablation experiments over a number of facets in order to better understand their relative importance, which is corresponding to Table 5-8. 

\textbf{In Table 5,} we show the ablation experiment results over in-domain pre-training corresponding to step 1 of Figure 1.  The gain of the text classification task (from 87.0\% to 88.2\%) shows the power of in-domain continued pre-training on unlabel data. Although the gain of NER task (from 88.7\% to 88.8\%) is small, we speedup the model inference by 200\%-300\% as we replace the 12-layer BERT to 3-layer BERT.

\textbf{In Table 6,} we show the ablation experiment results between unlabel pre-training and pseudo-label pre-training, corresponding to step 1 and 4 of Figure 1. The gain of the text classification task (from 91.7\% to 91.9\%)  proves that pseudo-label pre-training (i.e., the step 4 of Figure 1) can still improve the performance, even the fine-tuning dataset is very large.

\textbf{In Table 7,} we show the ablation experiment results between logits-based loss and one-hot label-based loss. For the task-specific pre-training step on text classification (i.e., the step 4 of Figure 1), using the KL-divergence loss with pre-softmax logits get more gain than the cross-entropy loss with one-hot pseudo-label. We think the reason is that the pre-softmax logits contains more pre-training information than one-hot pseudo-label.

\textbf{In Table 8,} we show the ablation experiment results between masked text and the origin text in the task-specific pre-training step. For the task-specific pre-training step (i.e., the step 4 of Figure 1), adding the MaskLM loss or masking 15\% tokens of the input text do not get more gain, compared to the no-mask or no-noisy text input. We also observer that no-mask or no-noisy text input for pre-training with pseudo-label is able to improve the performance, even when is fine-tuning dataset is relatively large (2000K).





\section{Conclusion}

In summary, we reveal these conclusions (detail conclusions are shown in Table 9-11):

1) For low-resource manually label dataset, the combination of all the steps of our learning framework improve the performance most, which means pre-training with no-masked logits-based pseudo-label data.

2) For high-resource manually labeled dataset, pre-training with task-specific loss (i.e., step 4 of Figure 1) without masking the input text is the only way to further improve the performance.

3) For pre-training with task-specific loss (i.e., step 4 of Figure 1), input the text without masking is better than masking 15\% tokens of the text.

4) For pre-training with task-specific loss (i.e., step 4 of Figure 1), using the KL-divergence loss computed on pre-softmax logits is better than cross-entropy loss computed on one-hot pseudo-label.

5) More labeled data diminishes the value of self-training. But pre-training still has big gain to performance.

6) Self-training as pre-training (i.e., step 4 of Figure 1) get bigger gain than in-domain pre-training (i.e., step 1 of Figure 1).

In the end, our main contribution is we answered the question: Combining pre-training and self-training to make best use of large amounts of unlabeled data improves the fine-tuning performance. We propose a learning framework using no-masked logits-based pseudo-label data for pre-training, which is superior to either pre-training or self-training alone. The experiment result shows that our learning framework make the best use of the unlabel data even when is fine-tuning dataset is relatively large. 


\bibliographystyle{ACM-Reference-Format}
\bibliography{sample-sigconf}


\end{document}